\documentclass{llncs}
\usepackage{multirow}
\usepackage{color}
\usepackage{graphicx}
\usepackage{tabularx}
\usepackage[export]{adjustbox}
\usepackage[caption = false]{subfig}
\usepackage{ marvosym }
\usepackage{siunitx}
\usepackage{amsmath}
\usepackage{cmap}
\usepackage{url}

\newcolumntype{C}{>{\centering\arraybackslash}X} % <-- modified
\graphicspath{{images/}}

\begin{document}
	\title{
		Improving Cytoarchitectonic Segmentation of Human Brain Areas with Self-supervised Siamese Networks
		\thanks{This is a pre-print of an article published in the Proceedings of the 21st International
		Conference on Medical Image Computing and Computer Assisted Interventions (MICCAI),
		Granada, Spain, September 2018.}
		%The final authenticated publication is available online at https://doi.org/[]
	}
	\author{
		Hannah Spitzer\inst{1}\textsuperscript{\Letter} \and 
		Kai Kiwitz\inst{2} \and
		Katrin Amunts\inst{1,2} \and 
		Stefan Harmeling\inst{3} \and 
		Timo Dickscheid\inst{1}
	}
	\institute{
		Institute of Neuroscience and Medicine INM-1, Forschungszentrum J\"ulich, Germany \\
		\Letter~\email{h.spitzer@fz-juelich.de} \and
		C. and O. Vogt Institute of Brain Research, Heinrich-Heine University D\"usseldorf, Germany \and
		Institute of Computer Science, Heinrich-Heine University D\"usseldorf, Germany
	}
	
	\maketitle
	
	\begin{abstract}
		Cytoarchitectonic parcellations of the human brain serve as anatomical references in multimodal atlas frameworks.
		They are based on analysis of cell-body stained histological sections and the identification of borders between brain areas.
		The de-facto standard involves a semi-automatic, reproducible border detection, but does not scale with high-throughput imaging in large series of sections at microscopical resolution.
		Automatic parcellation, however, is extremely challenging due to high variation in the data, and the need for a large field of view at microscopic resolution. 
		The performance of a recently proposed Convolutional Neural Network model that addresses this problem especially suffers from the naturally limited amount of expert annotations for training. 
		To circumvent this limitation, we propose to pre-train neural networks on a self-supervised auxiliary task, predicting the 3D distance between two patches sampled from the same brain. 
		Compared to a random initialization, fine-tuning from these networks results in significantly better segmentations. 
		We show that the self-supervised model has implicitly learned to distinguish several cortical brain areas -- a strong indicator that the proposed auxiliary task is appropriate for cytoarchitectonic mapping. 
	\end{abstract}
	
	\section{Introduction}
	Analysis of cytoarchitectonic cortical areas in high-resolution histological images of brain sections is essential for identifying the segregation of the human brain into cortical areas and nuclei \cite{amunts2015}.
	Areas can be distinguished based on their specific architecture, the presence of cell clusters and specific cell types according to their morphology, the visibility of columns, and other features. 
    Borders between cortical areas can be identified in a reproducible manner by a well-accepted method, which relies on image analysis and multivariate statistical tools to capture maximal changes in the distribution of cell bodies from the cortical surface to the white matter border \cite{schleicher1999}. 
    However, a completely automatic method that allows area segmentation in a large series of human brain sections is still missing. 
    
\begin{figure}[t]
		\centering
		\includegraphics[width=\textwidth]{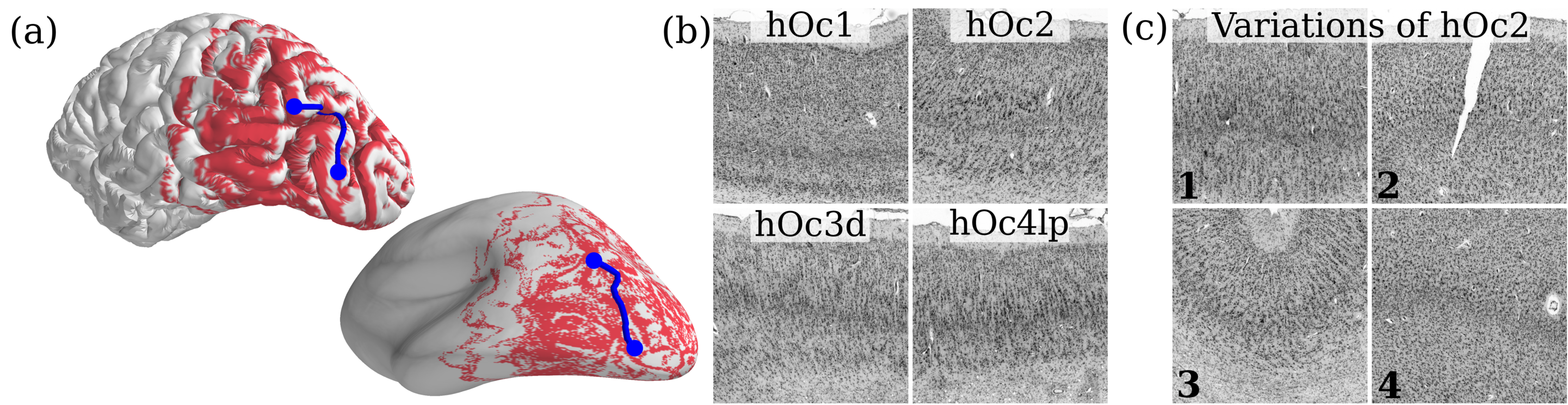}
		\caption{
			a) Sampling locations on pial and inflated left surface (red dots). 
			The approximate geodesic distance between two points is shown in blue.
			b) Example patches (1019$\times$\SI{1019}{px}) extracted from \SI{2}{\micro\meter} resolution histological sections showing areas hOc1--hOc4lp. Small variations in the laminar pattern distinguish the areas. 
			c) Examples of hOc2. 1) Intra-area variability, 2) artifacts, 3) high curvature, and 4) oblique histological cuts make identification of areas a challenging task.
		}
		\label{fig:sampling}
	\end{figure}

	Automatic identification of areas in cell-body stained whole brain sections is an extremely challenging segmentation task considering staining and sectioning artifacts, different relative orientations of the sectioning plane wrt.\ the brain surface, and high inter-subject variability (Fig.~\ref{fig:sampling}b, \ref{fig:sampling}c).
	To reliably identify differences in the distribution of cell bodies in the cortex, automatic methods need to rely on high resolutions (1--\SI{2}{\micro\meter}) and a large field of view (approx.\ \SI{2}{\milli\meter}) at the same time.
	Since expert annotation of brain regions is very labor-intensive, the amount of training data available for automatic algorithms is limited. 

	Previously, we have shown that despite these limitations, 
    it is possible to employ Convolutional Neural Networks (CNNs) for segmentation of cytoarchitectonic areas \cite{spitzer2017}. 
    However, the performance of this model is not yet accurate enough for fully automatic segmentation.
	
    In this work, we introduce a way to bypass the limitation of labeled training data by exploiting unlabeled high resolution cytoarchitectonic sections. 
    We formulate a self-supervised auxiliary task based on the estimation of spatial distances between image patches sampled from the same brain, using a Siamese network architecture.
    In particular, we determine the approximate geodesic distance between two image patches by exploiting the inherent 3D structure of a whole brain 3D reconstruction (Fig.~\ref{fig:sampling}a), as provided, e.g., by the \emph{BigBrain} \cite{bigbrain2013}.

	We make the following contributions: 
	1)~By applying transfer learning, we significantly improve the accuracy of area classification in the visual cortex. 
	2)~Carefully examining the training objective, we show that the Siamese network gains significant performance when predicting absolute 3D coordinates in addition to the pairwise distances.
	3)~We show that the self-supervised model learns to identify anatomically plausible cytoarchitectonic borders, although it was never trained to develop a concept of brain areas.

	\section{Related Work}
	\noindent\textbf{Automatic Area Parcellation.}
    Recently, CNNs were successfully employed for parcellation of MRT volumes  \cite{parc:brebisson2015,parc:glasser2016}. 
	In contrast to these works, we aim to improve cytoarchitectonic mapping in microscopic scans of human whole brain cell-body stained tissue sections.
	A CNN method for automatically classifying cortical areas in such data was introduced by \cite{spitzer2017}. 
	Building on a U-Net architecture \cite{ronneberger2015}, their key insight is to include prior knowledge in form of probabilistic atlas information to deal with the difficult variations in texture and limited training data. 
	In this work, we improve upon their results by leveraging unlabeled data in a self-supervised approach. 
    
    \medskip
    \noindent\textbf{Transfer Learning and Self-Supervised Learning.}
	In recent years, several studies have successfully used transfer learning to apply well performing models trained on the ImageNet dataset to new tasks \cite{trans:bengio2014}. 
	For the task at hand however, we require an unusually large receptive field and address a very specific data domain. 
	This forces us to train a custom CNN architecture from scratch.
	
	Siamese networks have been employed for learning highly nonlinear image features for keypoint matching, image retrieval, or object pose estimation \cite{siam:simo2015,siam:douman2016}. 
	Leveraging spatial dependencies in input images or exploiting motion information contained in video, such features can be learned in a self-supervised manner \cite{selfsup:doersch2015,selfsup:wang2015}. 
	We take up this idea and leverage the 3D relationship between individual brain sections.  
	In \cite{siam:douman2016} a Siamese regression network for pose estimation is presented, combining targets for pose regression from a single image with prediction of distance in pose between two input images. 
	Extending this approach, we include prediction of the 3D coordinates of input patches as an additional objective in our model, and explain the benefits gained by this modification.
	
	\section{Method}
	\begin{figure}[b]
		\centering
		\includegraphics[width=\textwidth]{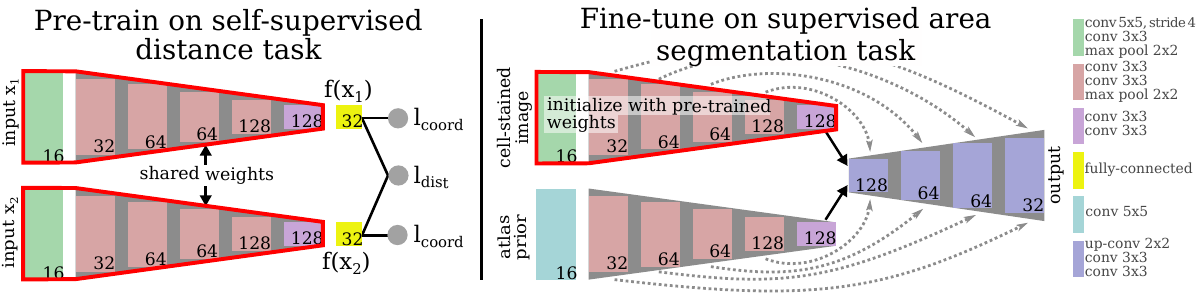}
		\caption{Siamese network architecture for the auxiliary distance task (left) and extended U-net architecture for the area segmentation task (right). The network branches marked in red share the same architecture. Based on \cite{spitzer2017}.}
		\label{fig:architecture}
	\end{figure}
	Our aim is to improve the accuracy of the supervised area segmentation in \cite{spitzer2017}. 
	Since classical cytoarchitectonic mapping is an extremely time consuming expert task, we cannot easily overcome the problem of limited training data. 
	We therefore propose to exploit unlabeled brain sections from a 3D reconstructed whole-brain volume for automatic brain mapping, of which much larger amounts can be acquired in reasonable time. 
	Our method consists of two consecutive steps (Fig.~\ref{fig:architecture}): 
	1) Pre-train weights
	on a self-supervised task using a Siamese network.
	2) Fine-tune 
	from these weights on the area segmentation task using a small dataset with brain region labels. 
	
	\subsection{Self-supervised Siamese Network on Auxiliary Distance Task}
	\label{sec:method:self}
	Considering a dataset of unlabeled brain sections from one human brain \cite{bigbrain2013}, we formulate the self-supervised feature learning task: 
	Given two input patches sampled randomly from the cortex in arbitrary sections, learn to predict the geodesic distance along the surface of the brain between these two patches (Fig.~\ref{fig:sampling}a). 
	
	We use a Siamese network that computes a regression function based on two input patches $(x_1, x_2)$ (Fig.~\ref{fig:architecture}).
	The network consists of two branches with identical CNN architecture and shared weights, computing features $(f(x_1), f(x_2))$. 
	The branch architecture corresponds to the texture filtering branch of the extended U-Net architecture of \cite{spitzer2017} with a 32-channel dense layer added on top of the last convolutional layer. 
	We define the predicted distance as the squared Euclidean distance between the feature vectors, and the distance loss as:
	\begin{equation}
	\label{eq:ldist}
	l_\text{dist} = \| \|f(x_1) - f(x_2)\|_2^2 - y_\text{dist} \|_1.
	\end{equation}
	The groundtruth distance $y_\text{dist}$ is computed by finding the closest points of the inputs on the brain surface and calculating their shortest distance along this surface.
	With this formulation of $l_\text{dist}$, the model learns a Euclidean feature embedding $f(x)$ of inputs $x$ wrt.\ the geodesic distance along the brain surface.
	
	We have successfully trained models using loss (\ref{eq:ldist}); however, our experiments have shown that convergence is faster and a higher accuracy regarding the predicted distances is reached, when we include the prediction of the 3D location of the inputs as an additional task in the training. 
	To this end we add an additional dense layer $d$ calculating the predicted coordinate for each input $x$ based on $f(x)$ and formulate the coordinate loss $l_\text{coord}$ as follows:
	\begin{equation}
	\label{eq:lcoord}
	l_\text{coord} = \|d(f(x)) - y_\text{coord}\|_1,
	\end{equation}
	with $y_\text{coord}$ the 3D location of input $x$ on the inflated surface.
	By defining $y_\text{coord}$ on the inflated surface, we ensure a high correlation between the distance of the coordinates and the geodesic distance of the inputs.
	For points on the right hemisphere, we reverse the left-right coordinate of $y_\text{coord}$ to account for the essentially mirror symmetric topology of areas on the two hemispheres.
	The coordinate loss $l_\text{coord}$ helps the network to learn a good feature embedding, agglomerating spatially close samples, even though they do not necessarily appear together as a pair during training.

	The final training loss is a weighted combination of $l_\text{dist}$ and $l_\text{coord}$ together with a L2 weight regularization that regularizes all weights and biases, except those in the final dense layers:
	\begin{equation}
	\label{eq:L}
	L = l_\text{dist}(x_1,x_2) + \alpha(l_\text{coord}(x_1) + l_\text{coord}(x_2)) + \lambda\|w\|_2.
	\end{equation}
	
	\subsubsection{Implementation Details}
	We generate our dataset from the \emph{BigBrain} \cite{bigbrain2013}, a dataset of $7400$ consecutive histological cell-body stained sections that were registered to a 3D volume at \SI{20}{\micro\meter} resolution.
	A surface mesh is available at \SI{200}{\micro\meter} resolution.
	We sample 200k 1019$\times$\SI{1019}{px} patches at \SI{2}{\micro\meter} resolution from sections 0--3000 (occipital and parietal lobe, encompasses visual cortex), leaving out $1/12$th of the sections for testing (cf.\ Fig.~\ref{fig:sampling}a for the sampling locations, Fig.~\ref{fig:sampling}b for example patches). 
	To ensure that the laminar structure of the cortex is clearly visible, we only sample from the center of non-oblique cortex, i.e., where the cutting plane was $\ang{90}\pm\ang{45}$ degrees to the brain surface.
 	From these samples we build 200k pairs in such a way that each patch occurs at least once, pairs always lie on the same hemisphere, and the resulting degree distribution of connections between pairs follows a power law. 
 	It would also be possible to include pairs across the hemispheres and calculate their distance by mapping points from the right hemisphere to the surface of the left hemisphere. We choose to stick to intrahemispheric distances due to interhemispheric differences in tissue size and area spread. 
	We set the coordinate loss weight $\alpha$ to 10 and the weight decay factor $\lambda$ to 0.001. Networks are trained for 16 epochs with SGD using an initial learning rate of 0.01, decaying by factor 2 every 3 epochs.
	
	\subsection{Fine-tuning the Extended U-Net on the Area Segmentation Task}
	For the area segmentation we use the extended U-net architecture proposed in \cite{spitzer2017}. 
	This model combines local image features extracted from high resolution image patches with a topological prior given by a cytoarchitectonic probabilistic atlas (\url{http://jubrain.fz-juelich.de}).
	For the two input types, the model has two separate downsampling branches that are joined before the upsampling branch.
	We use the same dataset as described in \cite{spitzer2017}, comprising 111 cell-body stained sections from 4 different brains, partially annotated with 13 visual areas using the observer-independent method \cite{schleicher1999}. 
	For training, 2025$\times$\SI{2025}{px} patches with \SI{2}{\micro\meter} resolution were randomly extracted from the dataset.
	
	We initialize the texture filtering branch with the weights from the Siamese distance regression network. 
	Compared to \cite{spitzer2017}, we reduce the overall learning rate, but train the atlas data branch with a higher learning rate to account for the different initializations of the branches.
	In detail, we first train for 8k iterations without the atlas data followed by additional 10k iterations including the atlas information (batch size 40). 
	Initial learning rates for these phases were 0.05 and 0.025 (0.25 for the atlas data branch), with learning rate decay at iterations 3k, 5k, and 6k by factor 2.
    Choosing a good learning rate was essential for the good performance of the fine-tuning. 

	\section{Experiments}
	We investigate the benefit of transfer learning from the self-supervised network on the original task of classifying brain areas. 
	In particular, we show the influence of the different loss components. 
	Furthermore, we demonstrate that the self-supervised network can distinguish several cytoarchitectonic areas without being explicitly trained on brain area classification.
	As evaluation metrics for the area segmentation task we report both the Dice score (harmonic mean of precision and recall) and the pixel distance error $err_\text{seg}$ that assigns to each misclassified pixel a penalty depending on the distance to the nearest pixel that is of the misclassified class \cite{yasnoff1977,spitzer2017}.
	For the self-supervised distance task the mean difference between the predicted and the groundtruth distances $err_\text{dist}$ is reported.

	\begin{figure}[tb]
	\subfloat{
		\begin{tabularx}{0.4\linewidth}{lCCC}
			\hline\noalign{\smallskip}
			\multirow{2}{*}{\shortstack[l]{Experiment\\(\#train, loss)}} & Aux. & \multicolumn{2}{c}{Area Segm.} \\
			&$err_\text{dist}$ & Dice & $err_\text{seg}$ \\
			\hline\noalign{\smallskip}
			Baseline \cite{spitzer2017}					  & -     & 0.72 & 21.2 \\ 
			\hline\noalign{\smallskip}
			20k, $l_\text{dist}$	    & 5.88 & 0.75 & 17.2 \\
			20k, $l_\text{coord}$ 	  & 8.13 & 0.79 & 16.8 \\
			20k, $L$ 				  & 4.54 & 0.79 & 15.2 \\
			\hline\noalign{\smallskip}
			\textbf{200k, $L$} & \textbf{3.73} & \textbf{0.80} & \textbf{14.4} \\
			\hline
		\end{tabularx}
	}
	\subfloat{
		\centering
		\includegraphics[width=.56\textwidth,valign=c]{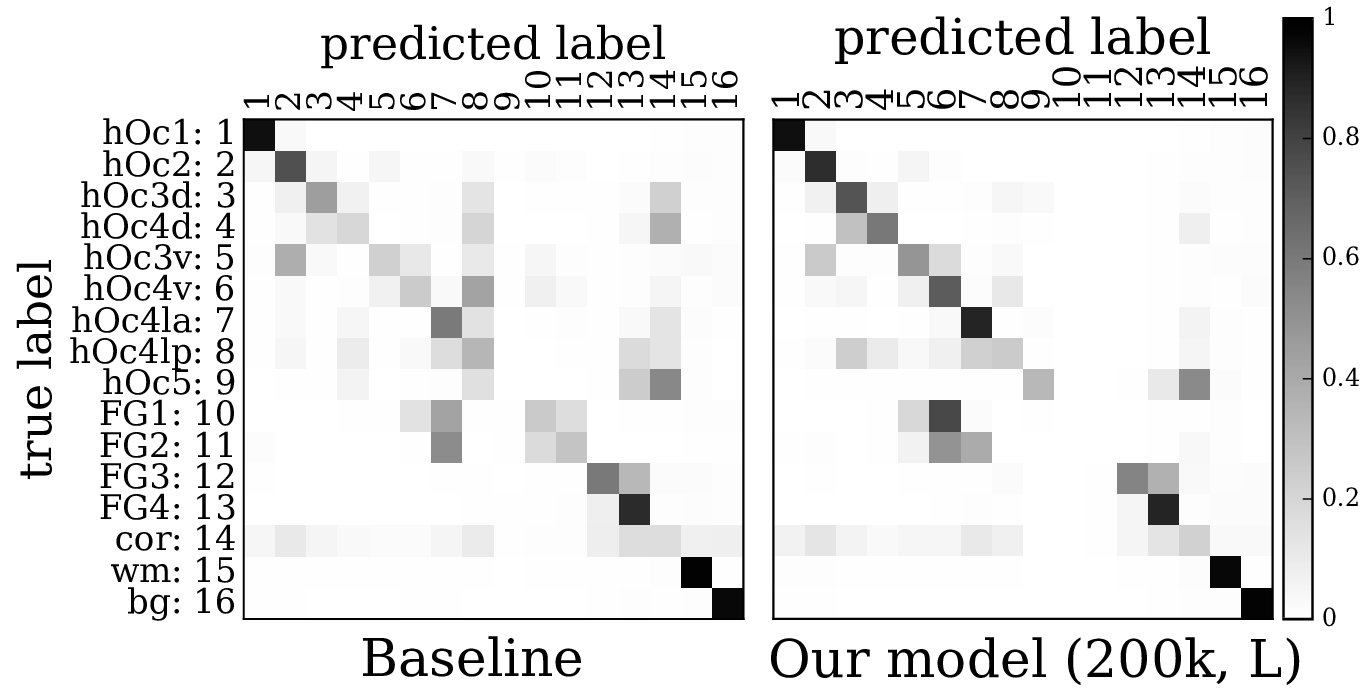}
	}
	\caption{
		Quantitative results.
		Left: Column 2 evaluates the models on the auxiliary distance regression task, and columns 3-4 show the performance of the fine-tuned models on the area segmentation task.
		Right: Confusion matrices for the baseline model and our fine-tuned model on the area segmentation task. Model were trained on visual areas hOc1--FG4, non-annotated other cortex, white matter, and background.
	}
	\label{fig:res}
	\end{figure}

	\begin{figure}[t]
	\centering
	\includegraphics[width=\textwidth]{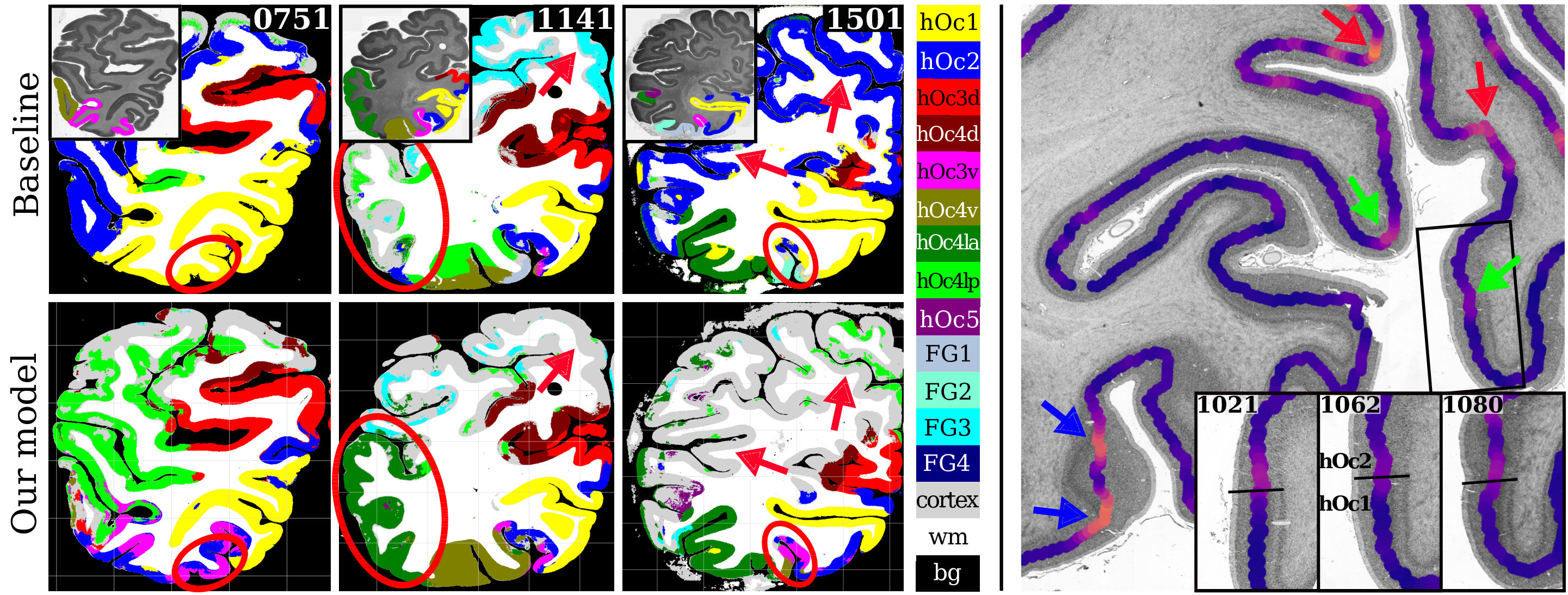}
	\caption{Qualitative results. 
		Left: Results on the area segmentation task with partial manual annotations in upper left corner. 
		Compared to the baseline \cite{spitzer2017}, our method predicts several areas significantly more accurate (circles) and has learned to deal much better with the ``other cortex" class (arrows). 
		Right: Squared Euclidean distances between averaged feature vectors of neighboring image patches, visualized by colored points along the cortical ribbon. 
		Blue values indicate lower distances than yellowish values. 
		Large distances occur at the border between hOc1/hOc2 over consecutive sections (green, enlarged boxes), at regions of high curvature (red) and at oblique regions (blue). This shows that the model has learned to recognize changes in cytoarchitecture.
		}
	\label{fig:eval}
	\end{figure}

	\medskip
	\noindent\textbf{Siamese Network Loss.}
	The loss function $L$ defined in Eq.~(\ref{eq:L}) for the self-supervised network combines a distance loss $l_\text{dist}$ with a coordinate loss $l_\text{coord}$. 
	In order to evaluate the influence of each loss component, we trained a self-supervised model on 10\% of the training set (20k samples) for 50 epochs.
	The model performs best when combining $l_\text{dist}$ with $l_\text{coord}$ (Fig.~\ref{fig:res}, rows 2-4 of the table). 
	The inclusion of $l_\text{coord}$ doubles the performance on the distance task, showing that $l_\text{coord}$ has the expected effect of guiding the model towards a more representative feature embedding. 
	When training only on $l_{coord}$ the performance of the fine-tuned network is almost as good as training on the combined loss. 
	However, the pixel distance error $err_\text{seg}$ is then larger. 
	A possible intuitive explanation is that $l_\text{dist}$ allows the model to see more realistic relationships between samples of the cortex than $l_\text{coord}$, where distances between coordinates only approximately represent the geodesic distance. 
	Thus the combined loss enables the model to better allocate individual samples in the feature embedding and make less errors on the area segmentation task. Training on the full dataset moderately increases performance on the area segmentation task.

	\medskip
	\noindent\textbf{Fine-tuned Area Segmentation Model.}
	Compared to the randomly initialized network in \cite{spitzer2017}, the Dice score increases to 0.80, while $err_\text{seg}$ drops from 21.2 to 14.4. 
	The drastic reduction of $err_\text{seg}$ indicates that due to the pre-training, the model can locate patches more accurately in the brain and less often confuses spatially distant areas. 
	Thus, the effect of pre-training on the distance regression task is similar to that of including the topological atlas prior in the supervised network \cite{spitzer2017}. 
	The confusion matrices (Fig.~\ref{fig:res}) and example segmentations (Fig.~\ref{fig:eval}) reveal that the fine-tuned model predicts more areas reliably, and overall exhibits less noise in the segmentation. 
	
	\medskip
	\noindent\textbf{Self-supervised Learning of Primary Visual Cortex.}
	To better understand the feature embedding that the self-supervised network learns, we average blocks of nine neighboring feature vectors (each \SI{200}{\micro\meter} apart) and plot the squared Euclidean distances between neighboring averaged feature vectors.
	This way we can appreciate the differences that the model predicts between neighboring regions.
	There are three main factors that cause the model to see large differences between neighboring parts of the cortex: 
	1) Oblique parts of the cortex, 2) regions with high curvature, and 3) borders between cortical brain areas.  
	The latter is particularly exciting: It shows us that the model actually discovered relevant properties of some cytoarchitectonic regions to solve the distance regression.
	In Fig. \ref{fig:eval} we show that the network has correctly identified the border between hOc1/hOc2 and tracks it through several sections. 
	
	\section{Discussion and Conclusion}
	Exploiting prior knowledge and the inherent structure of the data is beneficial for tasks with limited training data. 
	Our experiments show that the self-supervised distance task is a suitable auxiliary task for classifying cortical brain areas.
	It significantly increases the Dice score which is a measure for the quality of the segmentation. 
	In our evaluation, we have shown the importance of both components of our loss function to learn a good feature embedding.
	Additionally, we have demonstrated that our self-supervised model, trained with only the distances between samples as training signal, learns to identify several areal borders. 
	
	Inspired by this success, we plan to evaluate more auxiliary tasks based on inherent and relevant structures of 3D human brain reconstructions, such as local curvature or the relative orientation of the sectioning plane wrt.~the brain surface and further evaluate the unsupervised features and their applicability towards identifying areal borders. 
	
	\medskip
	\noindent\textbf{Acknowledgements.}
	{\footnotesize
	This work was partially supported by the Helmholtz Association through the Helmholtz Portfolio Theme “Supercomputing and Modeling for the Human Brain”, and by the European Union’s Horizon 2020 Framework Research and Innovation under Grant Agreement No. 7202070 (Human Brain Project SGA1).
	Computing time was granted by the John von Neumann Institute for Computing (NIC) and provided on the supercomputer JURECA at J\"ulich Supercomputing Centre (JSC).
    }
	
	%
	% ---- Bibliography ----
	%

\end{document}